\documentclass[accepted]{uai2025} 
                        
\usepackage[american]{babel}

\usepackage{natbib} 
\usepackage{mathtools} 
\usepackage{booktabs} 
\usepackage{tikz} 
\usepackage{hyperref}
\usepackage{pifont}
\usepackage{multirow}
\usepackage{makecell}
\usepackage{multicol}
\usepackage{appendix}
\usepackage[utf8]{inputenc}
\usepackage{marvosym}
\usepackage{amssymb}
\usepackage{footmisc}
\usepackage{authblk}

\usepackage{newfloat}
\usepackage{listings}
\DeclareCaptionStyle{ruled}{labelfont=normalfont,labelsep=colon,strut=off} 
\newcommand{\eg}{\textit{e.g., }} 
 
\newcommand{\etal}{\textit{et al.}} 
\newcommand{\etc}{\textit{etc}} 

\title{Creative Agents: Empowering Agents with Imagination for Creative Tasks}

\author[1]{Penglin Cai\thanks{Equal contribution.}}
\newcommand\CoAuthorMark{\footnotemark[\arabic{footnote}]}
\author[1]{Chi Zhang\protect\CoAuthorMark}
\author[1]{Yuhui Fu}
\author[1]{Haoqi Yuan}
\author[1]{Zongqing Lu\thanks{Correspondence to Zongqing Lu <zongqing.lu@pku.edu.cn>.}}
\affil[1]{%
    School of Computer Science\\
    Peking University\\
    Beijing, China
}

\begin{document}
\maketitle

\begin{abstract}
We study building embodied agents for open-ended creative tasks. While existing methods build instruction-following agents that can perform diverse open-ended tasks, none of them demonstrates creativity -- the ability to give novel and diverse solutions implicit in the language instructions. This limitation comes from their inability to convert abstract language instructions into concrete goals and perform long-horizon planning for such complicated goals. Given the observation that humans perform creative tasks with imagination, we propose a class of solutions, where the controller is enhanced with an imaginator generating detailed imaginations of task outcomes conditioned on language instructions.
We introduce several approaches to implementing the components of creative agents. We implement the imaginator with either a large language model for textual imagination or a diffusion model for visual imagination. The controller can either be a behavior-cloning policy or a pre-trained foundation model generating executable codes in the environment. We benchmark creative tasks with the challenging open-world game Minecraft, where the agents create diverse buildings given free-form language instructions. We propose novel evaluation metrics for open-ended creative tasks utilizing GPT-4V, which holds many advantages over existing metrics. We perform a detailed experimental analysis of creative agents, showing that creative agents are the first AI agents accomplishing diverse building creation in the survival mode of Minecraft. Our benchmark and models are open-source for future research on creative agents (\href{https://github.com/PKU-RL/Creative-Agents}{https://github.com/PKU-RL/Creative-Agents}). 
\end{abstract}

\section{Introduction}
\label{sec:intro}

\begin{figure*}[!t]
  \centering
  \includegraphics[width=.95\textwidth]{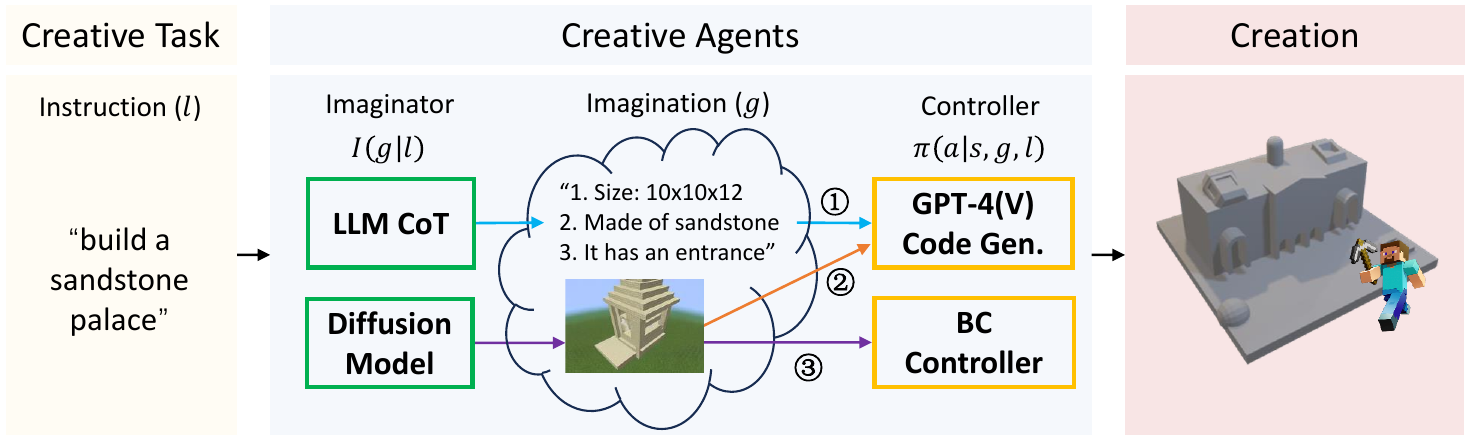}
  \caption{Overview of \textit{\textbf{creative agents}} for open-ended creative tasks. A creative agent consists of two components: an imaginator and a controller. Given a free-form language instruction describing the creative task, the imaginator first generates the imagination in the form of text/image by LLM with Chain-of-Thought (CoT)/diffusion model, then the controller fulfills the imagination by executing actions in the environment, leveraging the code generation capability of vision-language model (VLM) or a behavior-cloning (BC) policy learned from data. We implement three combinations of the imaginator and controller: (1) CoT+GPT-4, (2) Diffusion+GPT-4V, and (3) Diffusion+BC.}
  \label{fig:pipeline}
\end{figure*}

Building open-ended embodied agents has been a longstanding goal of AI research. Unlike many existing AI agents that perform a fixed set of tasks specified with rewards~\cite{drl-survey, dreamerv3}, open-ended agents can perform diverse arbitrary tasks without such specification. Existing research primarily focuses on learning instruction-following agents~\cite{saycan,steve1,VLP} that can solve open-ended tasks given free-form language instructions, achieving success in robotic domains~\cite{saycan,vima,VLP} and open-world games~\cite{gsb-mc, deps, steve1}. 
Among these, some agents can only follow clear instructions that represent specific goals or behaviors, and some other work focuses on task decomposition~\cite{ahn2022can, khot2022decomposed, li2022exploring, prasad2023adapt, tuli2022learning, vaezipoor2021ltl2action}. Creative tasks, where the instructions describe abstract tasks and the agent is required to generate complicated, novel, and diverse solutions, bring new challenges to intelligent agents.

As an example, in the open-world game Minecraft, existing agents can follow simple and clear instructions like \textit{`harvest a stone'}~\cite{plan4mc} and \textit{`build a snow golem, which stacks 2 snow blocks and 1 pumpkin'}~\cite{groot}, but they cannot solve creative tasks like \textit{`build a sandstone palace'}. For the latter, the agent can struggle to understand the target outcome of the task implied in the abstract instruction and plan actions for the long-horizon execution where hundreds of blocks should be properly placed. 
However, empowered with the ability of \textit{imagination}, humans can first imagine the appearance and functionality of the building, then plan for a proper order to build blocks and realize the imagined house in the game. Such ability enhances humans with strong creativity, enabling humans to create novel and diverse outcomes. Imagination also enriches the fuzzy instructions into refined task outcomes grounded in the environment, making the task description more explicit and executable.

Motivated by this ability, we introduce a framework for creative agents, empowering open-ended agents with imagination to solve creative tasks. Figure~\ref{fig:pipeline} gives an overview of the framework. 
Equipped with a text-conditioned imaginator, creative agents can imagine the details of the task outcome abstracted in the language instruction. These imaginations serve as a blueprint for the controller to interpret and act upon. We propose two variants of the imaginator, including a large language model (LLM)~\cite{gpt3} generating text imaginations and a finetuned diffusion model~\cite{stablediffusion} generating grounded visual imaginations.
We also introduce two variants of the controller that transform the imagination into executable plans. The first is a behavior-cloning controller trained on an environment dataset and maps imaginations to actions.
The second method leverages the strong abilities in vision-language understanding~\cite{yang2023dawn} and code generation~\cite{voyager} of the large vision-language model (VLM) GPT-4V~\cite{openai2023gpt}. The VLM controller receives the imagination as the task goal and generates code to perform actions in the environment. 

Designing evaluation metrics for open-ended tasks remains underexplored. Existing methods either use some surrogate metrics~\cite{voyager} which may not reflect the language instruction, or use human evaluation~\cite{groot} which is laborious. MineDojo~\cite{minedojo} proposes to use the similarity of the CLIP~\cite{clip} embedding between vision and language, which however can only provide some unknown correlation between the instruction and task outcome.  
To address these limitations, we propose novel evaluation metrics based on GPT-4V. Leveraging the analytical strength of GPT-4V, our metrics offer an effective, general, and human-independent means of evaluation. We verify that such metrics are consistent with human evaluations. Our proposed metrics are crucial for objectively measuring the creativity and effectiveness of solutions generated by open-ended agents. 

We benchmark creative tasks with challenging building creation in Minecraft\footnote{We select the open-world game Minecraft as the benchmark platform because it is convenient to build various imaginators and controllers and also supports creation in the game. Specifically, we choose the survival mode of Minecraft, where it is difficult for the agent to construct buildings since the agent has to move around and go up/down to place the blocks with diverse materials and colors, making the building process realistic. It is worth noting that our framework for creative agents is general and can also be applied to other environments.}, following 20 diverse instructions. Several variants of creative agents demonstrate their ability to create diverse and visually appealing buildings in the survival mode of Minecraft, which has never been achieved in previous studies. We give a detailed experimental analysis of creative agents, discuss the strengths and weaknesses of each variant, and provide insights for improving creative agents in future work.

\noindent {Our main contributions are threefold:}
\begin{itemize}
    \item We propose \textbf{\textit{creative agents}}, the first framework that endows open-ended agents with the ability to perform creative tasks through imagination. Our method builds the first instruction-following agent that can create diverse buildings in the survival mode of Minecraft.
    
    \item We establish novel evaluation metrics for creative tasks in open-ended environments, in which GPT-4V is used as the evaluator. 
    
    \item By open-sourcing the datasets and models, our work sets a new benchmark for future research in the field of open-ended learning and creative AI agents.
\end{itemize}

\section{Preliminaries}
\label{preliminaries}

\subsection{Open-Ended Tasks}

We formalize the process of the agent interacting with the environment as a Markov Decision Process (MDP) \textit{without reward}, defined by a tuple $M=(\mathcal{S}, \mathcal{A}, \mathcal{P}, \rho)$ representing states, actions, the transition function of the environment, and the initial state distribution, respectively. Starting from the initial state, for each time step, the agent performs an action based on the state, then the environment transitions to the next state upon the action.

Compared with traditional reinforcement learning tasks defined with reward functions, open-ended tasks have neither fixed targets nor optimal solutions. We follow MineDojo~\cite{minedojo}, formulating open-ended tasks as instruction-following problems $T=(\mathcal{L}, M)$, where $l\in\mathcal{L}$ is a free-form language instruction. We aim to acquire an instruction-following agent $P(a|s,l)$ which can exhibit behaviors consistent with the instruction to perform the described task.

\subsection{Creative Agents with Imagination}

Due to the abstract nature of language, language instructions cannot describe the full details of complicated tasks, drawing high uncertainty on the task completion and requiring the agent to possess creativity.
Though many open-ended agents~\cite{gsb-mc,steve1,VLP} can follow clear instructions that refer to some specific task goals, none of them can follow such uncertain instructions to perform complicated tasks.

We define creative tasks as a challenging case of open-ended tasks, where language instructions lack information to describe the whole task and can refer to diverse, novel, and complicated outcomes in the environment. Such instructions bring uncertainty for the agent and require the ability to imagine the details unspecified by the instruction. In addition, a short instruction (\eg `build a house') may refer to a long-horizon complicated task, increasing the challenge for the action planning and execution. 

To tackle the challenge, we propose to decompose the agent into an imaginator and a controller:
\begin{equation}
    P(a|s,l) = \sum_g{I(g|l)\pi(a|s,g,l)}.
\end{equation}
Here, $g\in G$ is an imagination of the task outcome, which can be in the form of diverse modalities (\eg text, image) and serves as a description of the target environment state of the task. The imaginator $I$ converts the instruction into an imagined outcome, providing the controller $\pi$ with a detailed task description. Therefore, we leave the uncertainty and creativity brought from creative tasks to the imaginator, providing the controller with richer task information to reduce its uncertainty. By disentangling these two models, we can delve deeper into the design choices for each part and combine them together to build creative agents.

\section{Generative Imagination}

Generative models in natural language processing and computer vision provide techniques to build the imaginator in either text space or image space.
In this section, we present two variants for implementing the imaginator.

\subsection{Language Models for Textual Imagination}

Large language models (LLMs) have shown marvelous abilities in solving diverse tasks~\cite{chang2023survey, wei2022emergent} as well as high plasticity with prompt engineering~\cite{gpt3, white2023prompt}. To tackle the problems in reasoning logically, Wei \etal~\cite{wei2022chain} proposed Chain-of-Thought (CoT), aimed at enhancing the emergence ability of LLMs.

Following the idea of zero-shot-CoT ~\cite{kojima2022large}, we design an imaginator using GPT-4~\cite{openai2023gpt} as the backbone, with zero-shot prompts for imagination in Minecraft building-creation domain (please refer to Appendix). 
Specifically, we provide the initial text instruction to GPT-4 and ask five questions relevant to the imagination, including the material used for the building, the approximate size, the significant features of the architecture, \etc. After GPT-4 generates answers to these questions indicating that the imagination process has been finished, we then ask the controller to execute actions accordingly to construct the building (see Section~\ref{sec:controller}). 

\subsection{Diffusion Models for Visual Imagination}

Diffusion models have achieved breakthrough performance in generating diverse and high-quality images. Stable Diffusion~\cite{stablediffusion} models data distribution as the stationary state of a diffusion process, learning to generate samples mirroring the true data distribution by reversing this process. Noteworthy for its training stability, it addresses issues like mode collapse.

To better align with the human conception of ``imagination'', we use images to be the imagination space and leverage text-conditioned diffusion models to be the imaginator. We finetune the Stable Diffusion~\cite{stablediffusion} using a text-image dataset to achieve a reasonable and diverse imagination of textual input. The text-image pairs in the dataset are constructed by automatically annotating the Minecraft buildings in CraftAssist~\cite{craftassist} using the multimodal Emu model~\cite{Emu}. 
After finetuning, we obtain visually plausible and diverse imaginations that align with both the textual descriptions and the Minecraft world.

\section{Designing Controllers}
\label{sec:controller}

After the imaginator generates the imagination $g$, it is the controller to take actions in the environment, conditioned on the current state, the imagination, and the language instruction. 
In the following, two variants of the controller are presented, including a behavior-cloning controller and a controller based on GPT-4(V). 

\subsection{Behavior-Cloning Controller}  

To transform the imagination into a practical construction process, we introduce a behavior-cloning controller that first converts the image imagination into a blueprint and then maps the blueprint into tangible actions. 

For tasks related to constructing buildings in Minecraft, we use voxel information as the basis for blueprints. To learn a module generating voxel blueprints conditioned on images, we adopt the methodology introduced by Pix2Vox++~\cite{pix2vox}, utilizing the image-voxel dataset constructed through data augmentation from original constructions in CraftAssist~\cite{craftassist} and ISM~\cite{ISM}. 
The module is trained to optimize a combination of the voxel prediction loss and two regularization terms, including the occupancy rate loss \cite{vpp} and the total variation loss \cite{voxurf,TV}.
Subsequently, for the construction process, we employ ResNet3D-CNN\cite{hara3dcnns} and train a behavior-cloning (BC) policy on a collected voxel-action dataset. After that, the final construction is executed by the BC policy conditioned on the voxel information through path-searching and block-placing within the MineDojo simulator~\cite{minedojo}. More details about our methods and datasets are available in Appendix.

\subsection{Vision-Language Models as Controller} 

We also adopt a generative vision-language model (VLM) to construct the controller, which can perceive both visual imaginations and textual imaginations. 
Utilizing its abilities in task reasoning and code generation, given an environment code interface that wraps actions, the VLM can generate executable code in the environment for task completion.

Specifically, we use GPT-4(V) which takes as input an image generated by the diffusion imaginator or the textual imagination generated by the LLM with CoT. We ask GPT-4(V) to generate code that can call Mineflayer~\cite{mineflayer} APIs to execute environment actions for building creation. Mineflayer implements JavaScript APIs for diverse skill primitives in the Minecraft world. 
Following the prompt design in Voyager~\cite{voyager}, we provide GPT-4(V) with API documentation to clarify the coding rules and a one-shot example of code generation for in-context learning. 
More details about the prompts are available in Appendix.

With this controller, we implement creative agents in both two modalities of imaginations.

\section{Experiments}

\subsection{Building Creation in Minecraft}

Inspired by the creative tasks in MineDojo~\cite{minedojo}, we set up an evaluation benchmark for constructing buildings in Minecraft, consisting of 20 diverse language instructions, such as ``a huge Minecraft volcano built of ice'' as illustrated in Figure~\ref{fig:showcase} and~\ref{fig:showcase_appendix}. Following the text description, the agent takes actions to move and place blocks in the game simulator to create buildings. 
In the experiment, we aim to investigate whether the agent can construct novel, diverse buildings by just following language instructions, which reflects the creativity of the agent. In the evaluation, we take screenshots of its creations in the game. More details can be found in Appendix~\ref{appendix:tasks}.
We setup various metrics to evaluate the open-ended building creation tasks and apply two evaluators, including human evaluators and a novel evaluator based on GPT-4V. Section~\ref{evaluation_metrics} presents the evaluation details.

\subsection{Implementation}


We implement several variants of creative agents using different combinations of imaginators and controllers (more details are provided in Appendix
) and build a baseline method to compare with: 

\begin{itemize}
\item {\bf Vanilla GPT-4.} This is the baseline method without imagination using GPT-4 as the controller. 
We simply use Voyager~\cite{voyager} with the original task instruction, and ask GPT-4 to perform code generation.

\item {\bf CoT+GPT-4.}  We implement this agent by adding a CoT-imagination on the basis of {\bf Vanilla GPT-4}, which means we use GPT-4 for both textual imagination and code generation (method (1) in Figure~\ref{fig:pipeline}).

\item {\bf Diffusion+GPT-4V\footnote{We use GPT-4V here to indicate the agent additionally takes an image imagination as input.}.} We use a finetuned Stable Diffusion to generate images as imagination and use GPT-4V as the controller to generate codes based on the visual imagination (method (2) in Figure~\ref{fig:pipeline}). 

\item {\bf Diffusion+BC.} The finetuned Stable Diffusion is used as the imaginator while the behavior-cloning controller is used to convert images into voxel blueprints and execute actions.
(method (3) in Figure~\ref{fig:pipeline}).

\end{itemize}

\subsection{Evaluation Metrics}
\label{evaluation_metrics}

Based on existing evaluation methods in open-ended learning \cite{mcu} and content generation in Minecraft \cite{gdmc2018}, we introduce a set of evaluation aspects, which are important criteria for creative agents:
\begin{itemize}
    \item \textbf{Correctness.} Are the creations consistent with the language instruction?
    \item \textbf{Complexity.} Can the agent create large and complex buildings?
    \item \textbf{Quality. } Do the creations have a good visual appearance from the perspective of aesthetics?
    \item \textbf{Functionality.} Do the created buildings have the necessary functions and structures (such as windows and entrances)?
\end{itemize}

It is worth noting that such evaluation metrics can directly reflect the agents' abilities of creating the buildings. In the meanwhile, such metrics also reach the same connotation of creativity - the ability to give novel and diverse task solutions implicit in the language instructions. In other words, in order to reach a higher level of performance, one agent must possess much more creativity.

To quantitatively evaluate such metrics, recent work \cite{groot} requires humans to perform evaluation, which however is labor-intensive and may be susceptible to subject preferences. 
To tackle these issues, we leverage the strong capabilities of the recent VLMs in vision-language reasoning and vision question answering and propose two VLM-based evaluation methods.

In the first method, given a language instruction, we sample a pair of creation results from two methods and fill in a template to ask the VLM which one is better overall based on all evaluation metrics:
\lstset{
     numbers = none
}
\begin{lstlisting}
You are a critic with high aesthetic ability. I will provide you a text instruction and two buildings created by different agents following this instruction in the game "Minecraft". 

Please evaluate their overall performance according to four aspects: $(EVALUATION_ASPECT 1~4). Tell me which building in the image is better (left or right).

Text: $(INSTRUCTION)
Image of buildings: $(IMAGE1, IMAGE2)
\end{lstlisting}

We use the Elo Rating System \cite{elo-rating} to measure the relative strength of each agent. 

In the second method, we fill in a template with the four evaluation metrics above to ask the VLM to directly score for each building. For each metric, the score is rated out of 10, and the overall score is the sum of the four metrics, which is out of 40.
\begin{lstlisting}
You are a critic with high aesthetic ability. I will provide you a text instruction and a created building following this instruction in the game "Minecraft".

According to four aspects: $(EVALUATION_ASPECT 1~4),
please evaluate the building with a score (out of 10) on each aspect respectively, then give a total score.

Text: $(INSTRUCTION)
Image of the building: $IMAGE
\end{lstlisting}

To verify the reliability of VLMs in evaluation, we also ask humans to participate in both two evaluation methods and compare the difference between VLM and human evaluations.  

\begin{figure}[!t]
  \centering
  \includegraphics[scale=0.28]{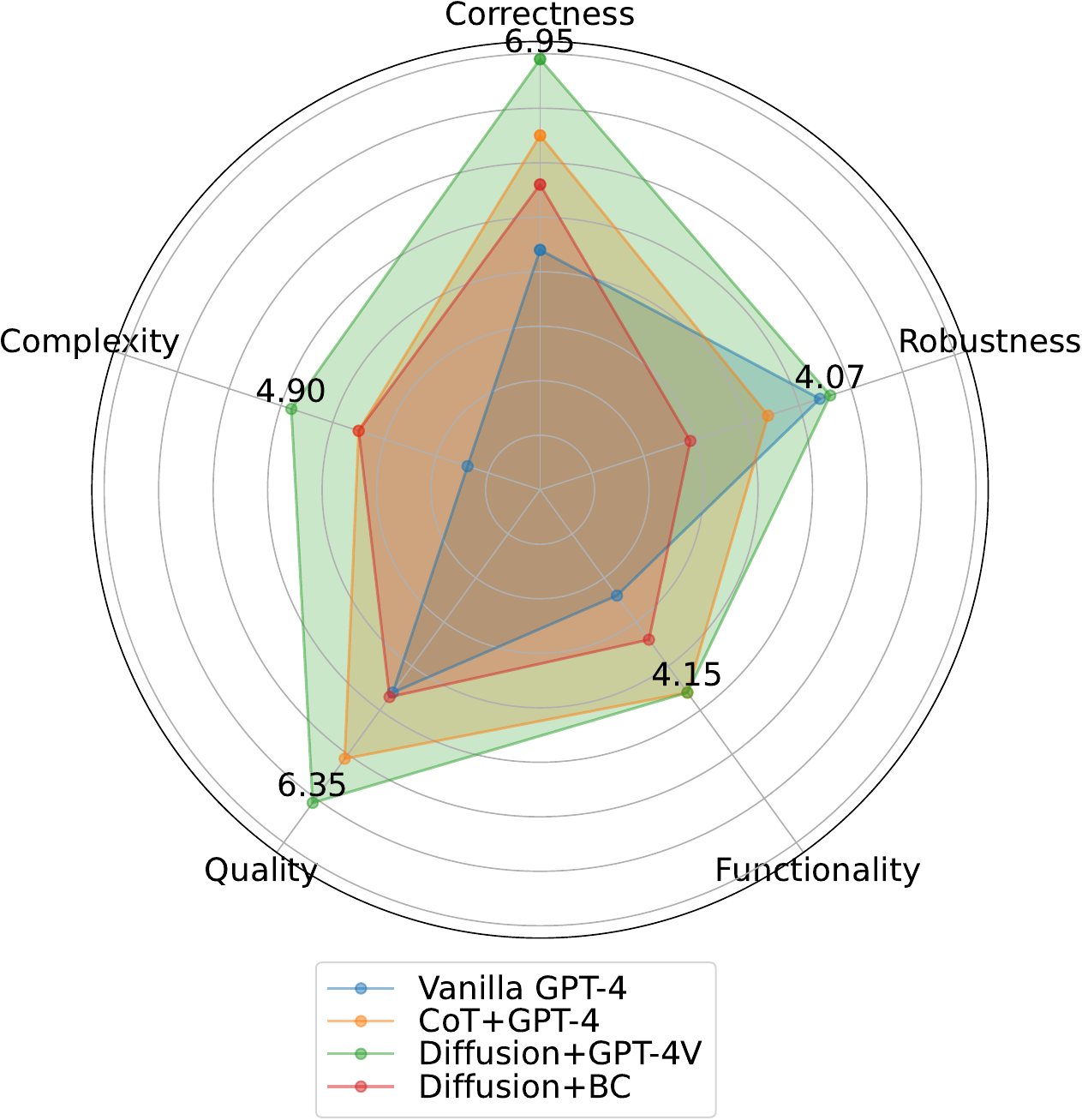}
  \caption{Comparison of all variants of creative agents in Minecraft building creation. For each evaluation metric, the number denotes the average score of the best agent over the 20 tasks. Diffusion+GPT-4V performs relatively better than other variants.}
  \label{fig:polygon}
\end{figure}

\begin{figure*}[!t]
  \centering
    \includegraphics[width=0.4\textwidth]{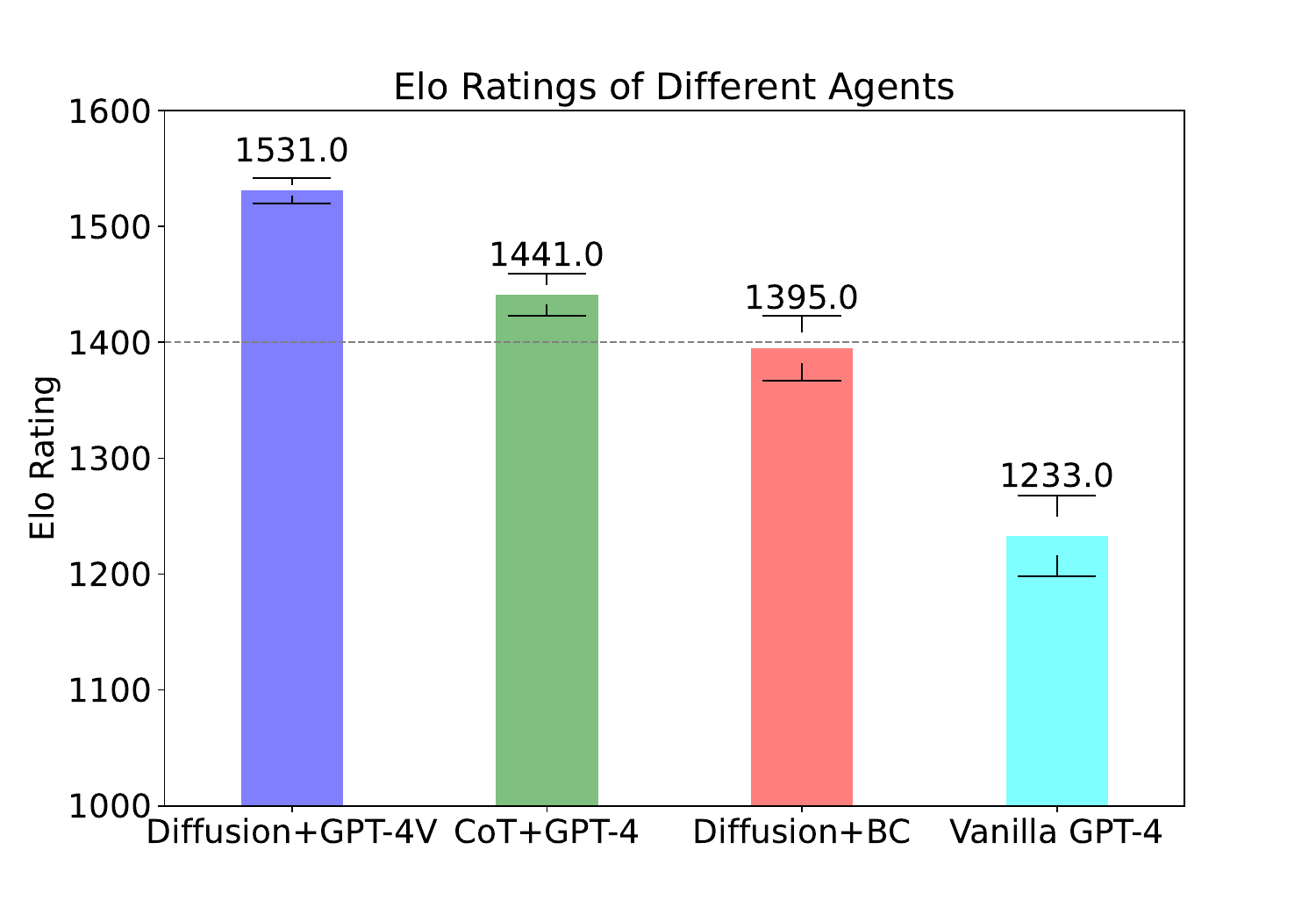}
    \includegraphics[width=0.4\textwidth]{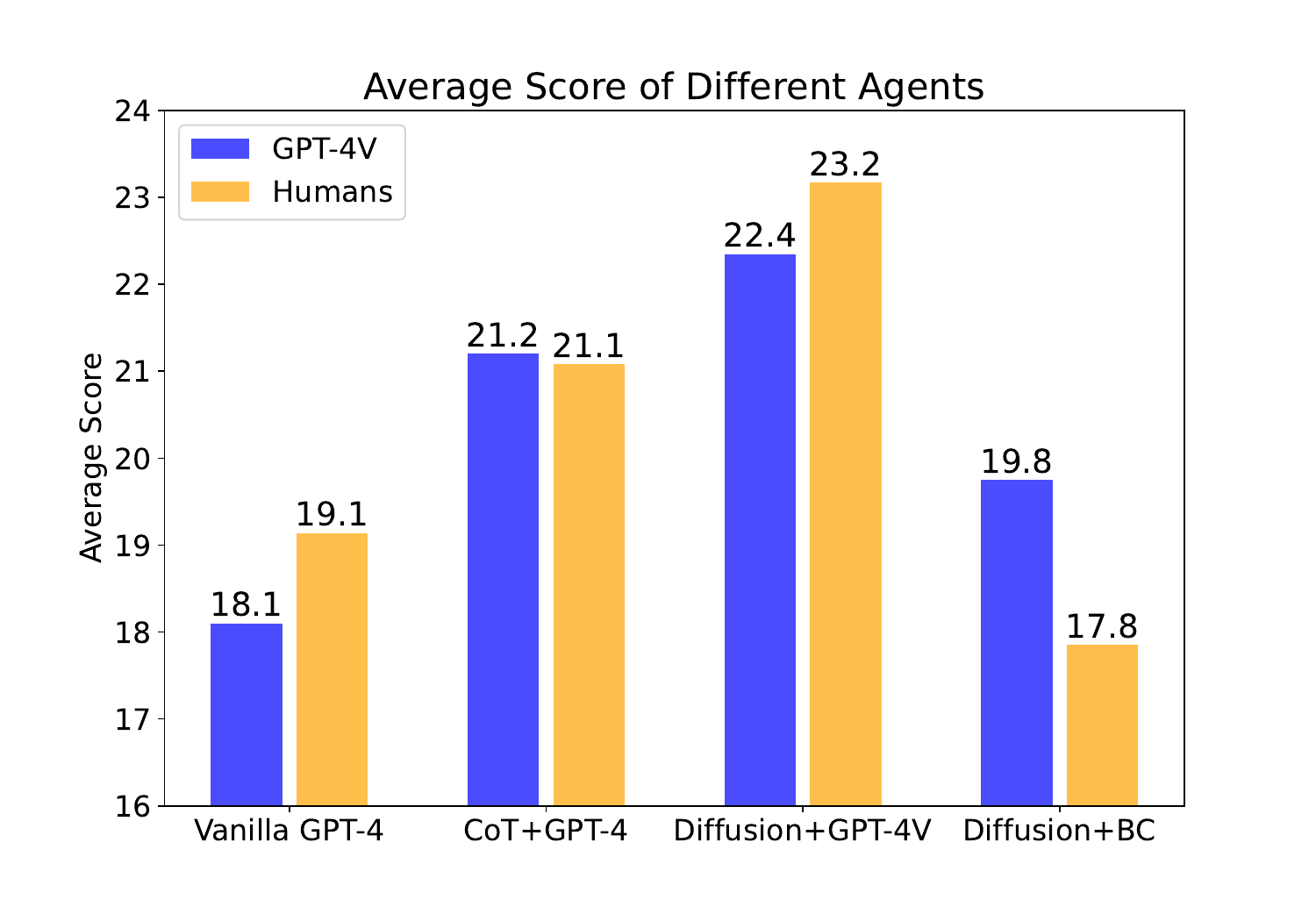}
  \caption{Evaluation results in Minecraft building creation. \textbf{Left}: The Elo Rating of all agents based on the evaluation of GPT-4V. \textbf{Right}: The average overall score of each agent in all test tasks evaluated by GPT-4V and humans.}
  \label{tab:results}
\end{figure*}

\subsection{Results and Analysis}

In the first evaluation method, every two agents are compared by GPT-4V for each test task. With all these comparison results, we model the four agents into a four-player zero-sum game and apply the Elo Rating System to compute the ratings, as shown in Figure~\ref{tab:results} (left).

According to the second evaluation method, we record ratings of each single test over the four metrics (correctness, complexity, quality, and functionality), the sum of which is the overall score. We calculate the standard deviation of the overall scores of all test tasks for each agent, which represents \textbf{Robustness}, with a larger standard deviation standing for weaker robustness. To align with the other four metrics for better presentation, we properly transform the standard deviation into a value such that a higher value indicates better robustness. Note that better robustness does not necessarily mean better performance since it merely indicates the consistency of the performance on all test tasks.
The results are plotted as a polar chart shown in Figure~\ref{fig:polygon}. Furthermore, the overall score averaged over all test tasks for each agent is shown in Figure~\ref{tab:results} (right). In Figure~\ref{fig:showcase} and~\ref{fig:showcase_appendix} in Appendix~\ref{appendix:examples}, we present qualitative results of the language description, the generated visual imagination, and the created building of each agent.

We also test the diversity using short language instructions with minimal information provided. The results are shown in Figure~\ref{fig:diversity} in Appendix~\ref{appendix:results}. With the same language instruction, the diffusion imaginator can generate diverse images for the controller, thus Diffusion+GPT-4V possesses a higher diversity than Vanilla GPT-4 and CoT+GPT-4.


Analyzing the experimental results, we can draw the following primal conclusions:

\begin{itemize}
    \item {\bf CoT for textual imagination can effectively enrich the detailed features of the target buildings.} 

    Comparing CoT+GPT-4 with Vanilla GPT-4, the former outperforms the latter in terms of all metrics except robustness in the polar chart by a large margin, and CoT+GPT-4 obtains a higher score in the Elo Rating results than Vanilla GPT-4. We assign this due to the rich information brought by Chain-of-Thought, which plays a role in self-reflection. Through this process, the LLM gets a better understanding of the details of the task, including but not limited to the materials used, the size of the building, the significant features, \etc. Within the context of a conversation, when GPT-4 generates the code in the second round, it can still perceive the extra information from the first round, thus reaching a better representation of the task goal.

    \item {\bf For the controller, using VLM instead of LLM leads to a marginally better performance in most metrics.}

    As shown in Figure~\ref{fig:polygon}, Diffusion+GPT-4V weakly surpasses CoT+GPT-4 in correctness, complexity, quality, and robustness. However, Diffusion+GPT-4V strikes a tie with CoT+GPT-4 in functionality.
    In terms of functionality, Diffusion+GPT-4V behaves no better than CoT+GPT-4, which can be owing to the weak ability of GPT-4V in 3D reconstruction. Empirically, the images passed to GPT-4V are usually a complete building generated by the diffusion-based imaginator, without the sectional view to show the internal structures. Therefore, sometimes GPT-4V tends to write code that leads to a solid house instead of a hollow one. According to the criteria of functionality, solid houses can result in low ratings.

    \item {\bf Diffsion+GPT-4V has the best performance overall, showing a strong ability of anti-interference and robustness.}

    In Figure~\ref{tab:results}, both the Elo Rating results and the average score show that the three variants proposed in Figure~\ref{fig:pipeline} outperform the baseline to varying degrees, among which Diffsion+GPT-4V ranks the first. Combining the previous two conclusions, Diffusion+GPT-4V has both the advantage in visual imagination and the strengths from CoT, thus having a better performance.
    Additionally, we are surprised to find that Diffusion+GPT-4V overcomes the misleading information of the diffusion-based imaginator. In about half of the test tasks, the images generated by the diffusion-based imaginator tend to have obvious noises in the background to some extent. However, GPT-4V seems to have the ability of anti-interference, thus capturing the major essential factors of the images. In contrast, Diffusion+BC may be susceptible to such noises, leading to weaker robustness. 

    \item {\bf The human-rating results coincide with the VLM-rating results with a minor gap, indicating that evaluating by vision-language models is reliable.}

    We list the average scores by both VLM and human evaluation in Figure~\ref{tab:results} (right), from which we know the human-rating results are generally in line with the VLM-rating results. In both evaluations, the first two in the ranking are the same - Diffusion+GPT-4V and CoT+GPT-4. The last two are in the opposite order but within a small gap. Overall, both two evaluations agree that Diffusion+GPT-4V has the best performance.


    \item {\bf The buildings created by agents are relatively simple, limited by the code written by language models and the trained policy of the behavior-cloning controller.}

    In the analysis of the final creations of different agents, we find that the buildings are relatively simple. For those variants with GPT-4(V) controllers, this may be limited by the code written by GPT-4(V). Due to limited APIs in Mineflayer, GPT-4(V) tends to generate simple code. For instance, GPT-4(V) tends to use for-loops in a piece of code that corresponds to a wall in the building, resulting in the square shape of the building. Additionally, Mineflayer uses a rule-based algorithm for path planning to place blocks in the Minecraft world, and the agent will always destroy some blocks when not able to find a proper path toward the goal. Therefore, there can be many demolished walls in the final creations. 
    On the other hand, the trained policy of the behavior-cloning controller has several limitations. When reconstructing voxels from the images generated by the diffusion-based imaginator, the Pix2Vox approach can only capture the RGB color for each voxel and choose the most similar block in Minecraft, which is not very accurate. To make things worse, the plausible structure of a common building is missed out during the reconstruction, which makes the voxel look like ``a mess''. Some blocks are even floating in the voxel, so they cannot be placed correctly in the final execution. This also provides a reason why Diffusion+BC ranks the last in the human evaluation results.
\end{itemize}

\begin{table}[ht]
    \centering
    \caption{Linear regression analysis and T-test across scoring metrics with statistical significance between GPT-4V and human's evaluation. Each row lists the coefficient, standard error, t-value, and p-value corresponding to the metric. For each scoring metric, the correlation coefficient between GPT-4V scores and human scores is significantly greater than 0, with p-values well below 0.05. Therefore, the positive correlation between them is statistically significant.}
    \label{tab:Ttest}
    \setlength{\tabcolsep}{3pt}
    \begin{tabular}{c|cccc}
        \toprule
        \textbf{Metric} & \textbf{ coef  } & \textbf{std err} & \textbf{   $t$   } & \textbf{ $P>|t|$ } \\
        \midrule
        Functionality & 1.0185 & 0.378 & 2.697 & 0.009 \\
        Quality & 0.9774 & 0.342 & 2.858 & 0.006 \\
        Complexity & 1.9722 & 0.281 & 7.021 & 0.001 \\
        Correctness & 1.4794 & 0.404 & 3.663 & 0.001 \\
        Total Score & 1.5910 & 0.315 & 5.048 & 0.001 \\
        \bottomrule
    \end{tabular}
\end{table}

\subsection{The VLM Evaluator vs. Human Evaluators}

In human evaluation, for the first evaluation method (1v1 comparison), we use the majority vote among all humans (49 human evaluators in total) to represent human preference for each pair of buildings.
For the second evaluation method, the scoring data from each human is standardized in each of the four evaluation metrics and the overall score. Then we take the average score from all humans as the human evaluation score for each building created by each agent. 

Based on the data of evaluation results of both GPT-4V and human, we conduct the linear regression analysis to test the relevant correlation between them. As shown in Table~\ref{tab:Ttest}, the results exhibit a high degree of correlation between the two methods of ratings, thus indicating that our method of evaluating via GPT-4V is reasonable and effective. For more details, please refer to Appendix~\ref{linear-regression}.

\section{Related Work}

\subsection{Open-Ended Agents}
In recent years, task learning in open-ended worlds has attracted increasing attention, among which Minecraft~\cite{johnson2016malmo} has become a marvelous test-bed for open-ended learning. MineRL \cite{minerl} and MineDojo \cite{minedojo} implemented simulated environments and organized datasets of a relatively large scale, and the latter provides tasks for agent training in the open-ended domain. However, most previous work~\cite{vpt, deps, plan4mc, steve1, ptgm} mainly focused on unlocking numerous skills and tackling long-horizon tasks in Minecraft, which however are mostly predefined, lacking the open-ended nature, not to mention creativity. 

The IGLU Competition~\cite{kiseleva2022interactive} was a giant leap to solving tasks according to instructions in natural language. Skrynnik \etal~\cite{embodied2} proposed a pipeline containing a T5-based language module, a rule-based transfer module, and the downstream policy module for execution. This agent could solve simple tasks of stacking blocks in Gridworld~\cite{zholus2022iglu}, a Minecraft-like open-ended world. However, it depended on too many step-by-step instructions, thus showing little creativity. In general, there is a significant gap between previous work and the true ``creative agents''.

\subsection{Generative Models} 
In recent years, many modern generative models have been proposed and are used to produce high-quality samples in various domains \cite{cao2022survey}.
Text-generative models have aroused much attention due to their wide range of uses. Especially, large language models (LLMs) are playing more and more significant roles in decision-making, planning, and reasoning. Among LLMs, a representative one is GPT-4~\cite{openai2023gpt}, whose emergence has laid a solid foundation for further research. Accompanied by the appearance of LLMs, prompt engineering and tuning techniques~\cite{gpt3, white2023prompt} have been widely studied and applied, including Parameter-Efficient Fine-Tuning (PEFT)~\cite{he2021towards} and Chain-of-Thought (CoT)~\cite{wei2022chain}.
In our work, LLMs with CoT are adopted as textual imaginators, and we also construct the text-based controller with LLM code generation.

In the field of computer vision, image-generative models are becoming increasingly important. 
Prominent approaches include variational autoencoders (VAEs)~\cite{VAE}, Generative Adversarial Networks (GANs)~\cite{GAN}, and flows~\cite{flow++}, demonstrating success in capturing image distributions and representations.
Recently, diffusion models \cite{diffusion} and DALL-E 3~\cite{dalle3} are springing up, accelerating the research in visual generation. In our work, a finetuned Stable Diffusion~\cite{stablediffusion} is used for visual imagination, representing a concrete description of the building-creation task.

In the Minecraft world, previous work~\cite{gdmc2018, world-gan, evocraft} focuses on Procedural Content Generation (PCG). However, they usually generate a pre-defined type of buildings with a lot of human prior knowledge. 
In our work, imagination for creative agents is similar to content generation, but our imaginator can generate with free-form instructions, in different modalities, and requiring much less human prior.

\subsection{Evaluation for Open-Ended Tasks}
Recent work has gathered many evaluation methods for open-ended tasks. Voyager~\cite{voyager}, STEVE-1~\cite{steve1}, and DIP-RL~\cite{dip-rl} use travel distance and collected items as surrogate metrics to evaluate. GROOT~\cite{groot} and BASALT competition~\cite{BASALT} use human evaluation, which is relatively labor-intensive and may be susceptible to subject preferences. Recent work~\cite{minedojo,clip4mc} proposes to use the CLIP-like model to compute alignment between the behaviors and instructions. We propose a novel evaluation method using VLMs, which can either directly rate the performance in various aspects or conduct pairwise comparisons.

In terms of evaluation aspects, previous studies have proposed a variety of metrics. MCU~\cite{mcu} took evaluations from the perspective of planning complexity, time consumption, novelty, and creativity. GDMC Competition~\cite{gdmc2018} required humans as judges, rating the generated contents from adaptability, functionality, evocative narrative, as well as visual aesthetics. Stooke \etal~\cite{open-ended-learning} evaluated the results in both task coverage and relative performance. Inspired by these studies, we adopt the evaluation aspects in correctness, complexity, quality, functionality, and robustness.

\section{Conclusion and Limitations}

In this paper, we propose \textbf{\textit{creative agents}}, which is the first framework that can handle creative tasks in an open-ended world. Using this framework, we implement various embodied agents through different combinations of imaginators and controllers. Additionally, we tap into the potential of Vision-Language Models (VLMs), utilizing VLMs for evaluation as judges. By comparing the rating results from VLM and humans, we illustrate the reliability of VLM evaluation. 

In the meanwhile, we find a few limitations of these creative agents, to be investigated in further work. First, there is much room for improving the BC controller, especially for the performance of Pix2Vox module. Another limitation lies in the simplicity of the building created by the agents, which means the capabilities of these agents are limited. How to enhance the creativity of agents can be a challenging problem.

In the end, we declare that \textbf{\textit{creative agents}} is an initial attempt in this field, aimed at raising the awareness of building intelligent agents with creativity. We hope this work can be of inspiration for further research.

\section*{Acknowledgments}

This work was supported by NSFC under Grant 62450001. The authors would like to thank the anonymous reviewers for their valuable comments and advice.

\bibliography{uai2025-template}

\newpage

\onecolumn

\title{Creative Agents: Empowering Agents with Imagination for Creative Tasks\\(Supplementary Material)}
\maketitle

\appendix

\section{Qualitative Results}
\label{appendix:examples}
\begin{figure*}[htbp]
  \centering
  \includegraphics[width=.96\textwidth]{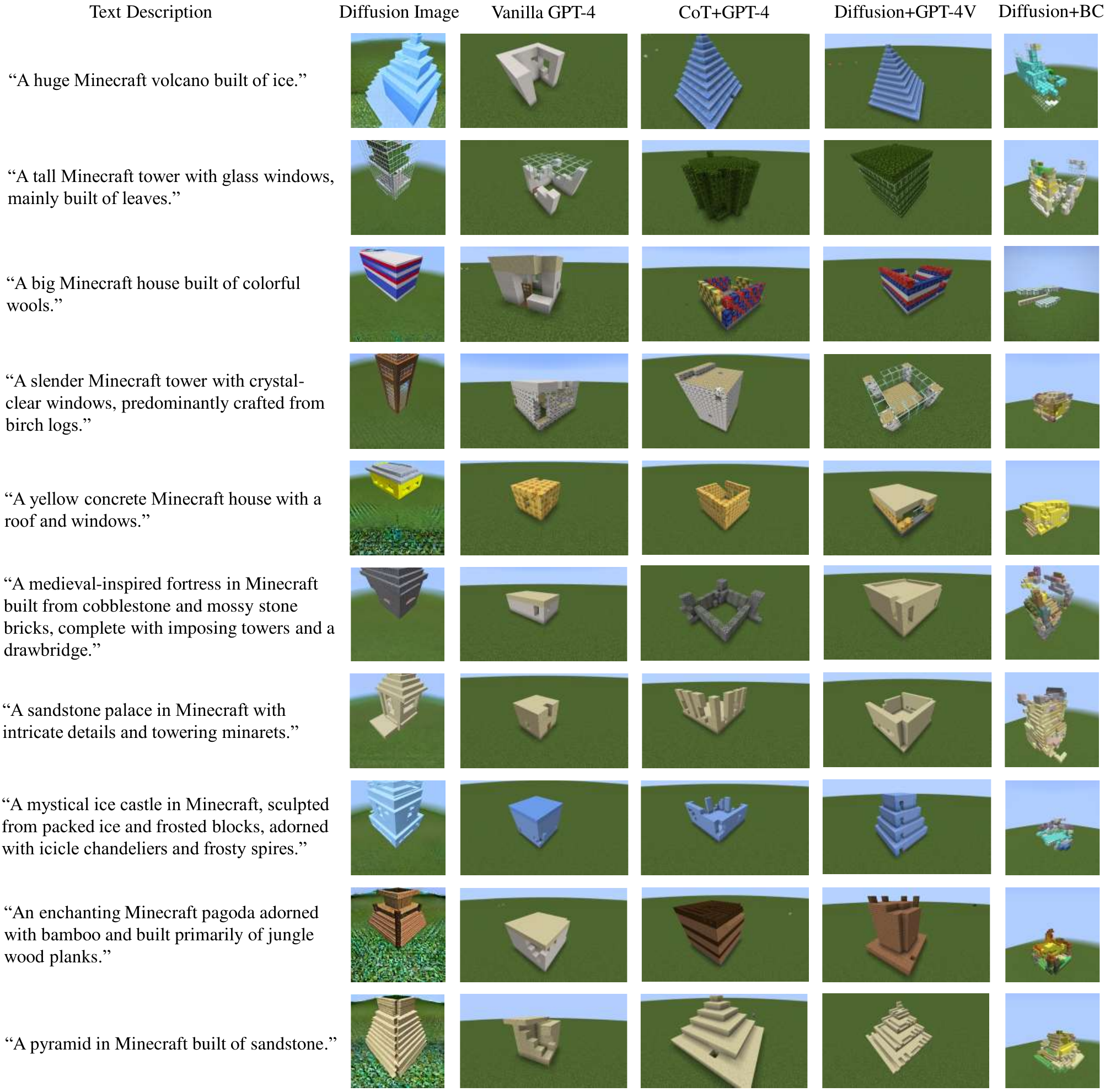}
  \caption{Qualitative results of the language description, the generated visual imagination,
and the created building of each variant of creative agents. Visual imagination generated by the diffusion model has great diversity, which is an important manifestation of creativity.}
  \label{fig:showcase}
\end{figure*}

\begin{figure*}[!t]
  \centering
  \includegraphics[width=.96\textwidth]{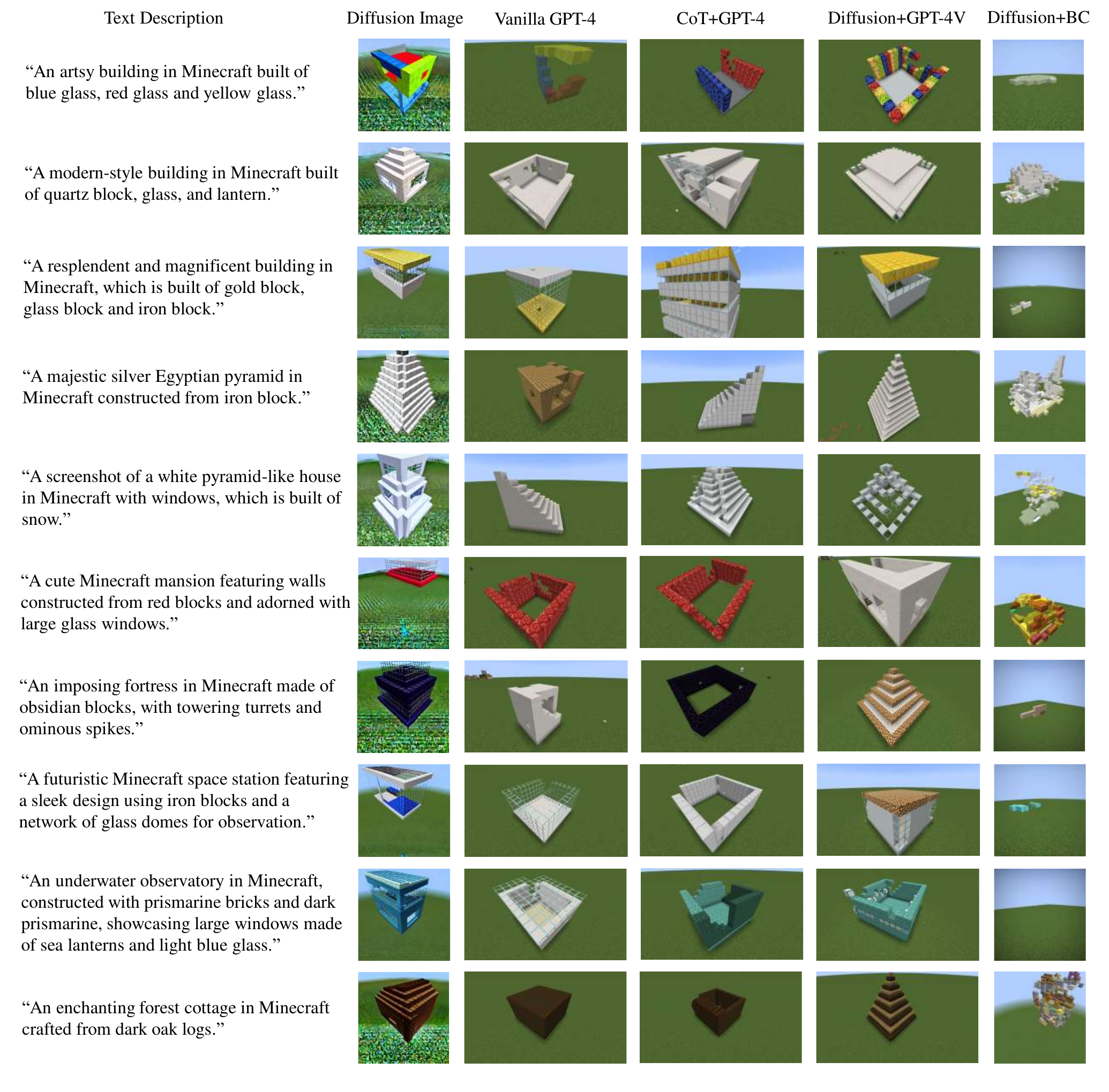}
  \caption{Additional results of the language description, the generated visual imagination, and the created building of each variant of creative agents.}
  \label{fig:showcase_appendix}
\end{figure*}

In this section, we present the showcases of the results of building creation (Figure~\ref{fig:showcase} and~\ref{fig:showcase_appendix}), including the language instruction, the generated visual imagination, and the created building of each variant of creative agents.

\section{Benchmark Details}
\label{appendix:tasks}

In this section, we present our Minecraft Building Creation benchmark for creative agents, including tasks, environment simulators, and datasets.

We introduce a set of language instructions for test tasks. Each instruction is constructed by randomly picking a building name and several building features and combining them into a natural language description. We list the 20 instructions here:
\begin{lstlisting}
    1. An artsy building in Minecraft built of blue_glass, red_glass and yellow_glass.

    2. A pyramid in Minecraft built of sandstone.

    3. A modern-style building in Minecraft built of quartz_block, glass, and lantern.

    4. A resplendent and magnificent building in Minecraft, which is built of gold_block, glass_block and iron_block.

    5. A majestic silver Egyptian pyramid in Minecraft constructed from iron block.

    6. A screenshot of a white pyrmaid-like house in Minecraft with windows, which is built of snow.

    7. A tall Minecraft tower with glass windows, mainly built of leaves.

    8. A cute Minecraft mansion featuring walls constructed from red blocks and adorned with large glass windows.

    9. A big Minecraft house built of colorful wools.

    10. A huge Minecraft volcano built of ice.

    11. A slender Minecraft tower with crystal-clear windows, predominantly crafted from birch logs.

    12. An imposing fortress in Minecraft made of obsidian blocks, with towering turrets and ominous spikes.

    13. A futuristic Minecraft space station featuring a sleek design using iron blocks and a network of glass domes for observation.

    14. An underwater observatory in Minecraft, constructed with prismarine bricks and dark_prismarine, showcasing large windows made of sea lanterns and light_blue_glass.

    15. An enchanting Minecraft pagoda adorned with bamboo and built primarily of jungle wood planks.

    16. A sandstone palace in Minecraft with intricate details and towering minarets.

    17. A mystical ice castle in Minecraft, sculpted from packed ice and frosted blocks, adorned with icicle chandeliers and frosty spires.

    18. An enchanting forest cottage in Minecraft crafted from dark oak logs.

    19. A yellow concrete Minecraft house with a roof and windows.

    20. A medieval-inspired fortress in Minecraft built from cobblestone and mossy stone bricks, complete with imposing towers and a drawbridge.
\end{lstlisting}

We adopt two environments for different creative agents, each providing a Minecraft simulator with action primitives for the agent to place a block.

The first environment uses the MineDojo simulator~\cite{minedojo}, which provides Gym wrappers for the Minecraft game along with additional dictionary observations. We implement two primitive actions in this simulator: a path-search action and a placing action. 
The path-search action, implemented with the A* algorithm, can use voxel information to determine whether the agent can reach a given coordinate. The placing action can drive the agent towards a target coordinate given a valid path and place a block on it.
A creative agent can recursively output the plan for the next block to place and call the action primitives to accomplish it.

In the second environment, we use the original Minecraft game (Java Edition 1.19). 
We use Mineflayer~\cite{mineflayer} with JavaScript APIs to provide action primitives for building creation. 
The core function we use is \textit{placeItem(bot, block name, position)}, which also performs path-search and placing to try to place a block at a given position.

The first environment with MineDojo simulator has the advantage of providing richer observation information, which is beneficial for data collection and programmatic evaluation in future research. The second environment with Mineflayer APIs provides more diverse action primitives, which may be helpful for future research on other creative tasks. In our study, we test the method with BC controller in the first environment and test all other methods in the second environment.

For datasets, we release a text-image dataset for training diffusion imaginators, an image-voxel dataset, and a gameplay dataset for training the BC controller.
The text-image dataset consists of 14,180 paired language instructions and images. Each RGB image has a resolution of $512\times512$. 
The image-voxel dataset comprises 1,009,044 paired images and 3D voxels. Each $512\times512$ image shows a Minecraft building at a view angle. Each $32\times32\times32$ voxel is the corresponding ground-truth voxel of the building, labeled with voxel occupancy and block information. 
The gameplay dataset consists of 6M action-labeled samples. Each trajectory $(V, \{a_t\}_{t=0}^{T})$ is an expert gameplay to construct a building, where $V$ denotes the voxel of the target building and $\{a\}$ denotes the sequence of action primitives to complete it. 

\section{Implementation Details}
\label{appendix:implement}

In this section, we present details of data collection and implementing imaginator and controller variants.

\subsection{LLM Prompts for Textual Imagination}

To utilize GPT-4 as a textual imaginator and make it have a better understanding of the building task, we add Chain-of-Thought (CoT) prompts in addition to the prompts for \textbf{Vanilla GPT-4}, following Wei \etal~\cite{wei2022chain}. Specifically, we ask GPT-4 five questions in the first round within a conversation, using the following template:

\begin{lstlisting}
    You are an architect designing houses and buildings.
    Here is a building you should design: _______________________________.
    
    First, you should answer these questions below based on your design and imagination:
    
    1. Please give a detailed description of the building.
    2. Which kinds of blocks are used in the building? There might be several kinds of materials, and you should report them all.
    3. What is the probable size of the building? You should estimate the length, width and height. Note that your inventory contains only 2304 blocks, so length, width and height no more than 12 would be better.
    4. Please list some components of the building, including but not limited to doors, walls, windows and floors.
    5. Are there any salient features of the building, e.g. gardens, swimming pools and towers?
\end{lstlisting}

After GPT-4 answers these questions as imagination, in the second round of the conversation, we ask it to generate code to construct the building. The prompts for the second round are presented in Appendix~\ref{VLM Controller}.

\subsection{Finetuning the Diffusion Model for Visual Imagination}

To finetune the diffusion model for generating visual imaginations, we construct the text-image dataset by automatically annotating the Minecraft buildings in CraftAssist~\cite{craftassist} using the multimodal Emu model~\cite{Emu}. The labeled dataset consists of 14K text-image pairs. 

In order to obtain stable and high-quality image generation with such a small dataset, we choose to finetune the pre-trained StableDiffusion V1.4 model~\cite{stablediffusion}. Table~\ref{tab:diffusion_hp} shows the hyperparameters used for finetuning. 

\begin{table}[htbp]
  \centering
  \caption{Hyperparameters for finetuning Stable Diffusion.}
  \setlength{\tabcolsep}{3pt}
  \begin{tabular}{cc}
    \toprule
    Hyperparameter & Value \\
    \midrule
    Training epoch & 100 \\
    Image resolution & $512\times 512$ \\
    Adam $\beta$ & $(0.9,0.999)$ \\
    Adam weight decay & 0.01 \\
    Learning rate & $1\times 10^{-4}$ \\
    Max grad norm & 1.0 \\
    \bottomrule
  \end{tabular}
  \label{tab:diffusion_hp}
\end{table}

\subsection{Behavior-Cloning Controller}

For the behavior-cloning (BC) Controller, we utilize the voxel representation as a blueprint. We transform images into voxels, and subsequently convert the voxel blueprint into a sequence of actions. 

For data collection, we perform data augmentation on the original voxel data obtained from CraftAssist~\cite{craftassist} and use MineDojo~\cite{minedojo} to construct the buildings and render the images. We collect 1M paired images and voxels to construct the dataset and split a validation set with 3K samples.
We employ the Pix2Vox++ architecture~\cite{pix2vox}, with an encoder of 30M parameters and a decoder of 70M parameters. The model hyperparameters are shown in Table~\ref{tab:pix2vox_hp}. It resizes the input image into 224$\times$224 pixels and generates a 32$\times$32$\times$32 voxel output. 
Each voxel has 4 dimensions, consisting of the probability of occupancy and RGB colors for the block.

\begin{table}[htbp]
  \centering
  \caption{Hyperparameters for Pix2Vox++.}
  \setlength{\tabcolsep}{3pt}
  \begin{tabular}{cc}
    \toprule
    Hyperparameter & Value \\
    \midrule
    Encoder layers & [3, 5, 5, 3] \\
    Encoder block inplanes & [64, 128, 256, 512] \\
    Decoder layers & [1, 1, 1, 1, 1] \\
    Decoder block inplanes & [2048, 512, 128, 32, 8] \\
    Image resolution & $224\times 224$ \\
    Output shape & $32\times 32\times 32 \times 4$ \\
    Learning rate & $1 \times 10^{-3}$ \\
    Weight decay & $1\times 10^{-4}$ \\
    \bottomrule
  \end{tabular}
  \label{tab:pix2vox_hp}
\end{table}

For generating action plans given voxels, we collect a gameplay dataset for building creation and training a behavior-cloning policy.
We use voxels in the image-voxel dataset to construct diverse goals for building and collect a dataset of 6M steps. 
We preprocess the dataset to provide rich observations for the BC policy.
We construct the observation at each timestep with $32\times 32\times 32\times 3$ voxels. Each channel represents the target voxel, the built voxel at the current step, and the last block, respectively. 
The output shape is $32769=32\times 32\times 32 + 1$, representing the action primitive for the position of the next block to place and the termination probability.
We adopt the ResNet3D~\cite{hara3dcnns} architecture with 50M parameters. The hyperparameters are shown in Table~\ref{tab:planning_model_hp}.

\begin{table}[htbp]
  \centering
  \caption{Hyperparameters for the BC policy.}
  \setlength{\tabcolsep}{3pt}
  \begin{tabular}{cc}
    \toprule
    Hyperparameter & Value \\
    \midrule
    ResBlock Type & BasicBlock \\
    Layers & [2, 2, 2, 2] \\
    Block inplanes & [64, 128, 256, 512] \\
    Input shape & $32\times 32\times 32\times 3$ \\
    Output shape & 32769 \\
    Learning rate & $1 \times 10^{-3}$ \\
    Weight decay & $1\times 10^{-4}$ \\
    \bottomrule
  \end{tabular}
  \label{tab:planning_model_hp}
\end{table}


\subsection{VLM Controller}
\label{VLM Controller}


Voyager~\cite{voyager} uses Mineflayer APIs as action primitives to solve tasks in the Minecraft world. For building creation tasks, we mainly utilize the function \texttt{position.offset(x, y, z)} to convert planned coordinates into positions and use \texttt{placeItem(bot, blockName, targetPosition)} to place a block at the target position.

After textual imagination or visual imagination in the first round, we ask GPT-4(V) to generate executable code in the environment according to the imagination and context.
The prompts mainly consist of three parts:
(1) Mineflayer APIs and their usage.
(2) Explanations on the coding format, the task, etc. 
(3) An example of the correct code for in-context learning.

We present the prompts used in the second round of the conversation here.

\begin{lstlisting}
    Based on (the image and) your answers to the questions above, please design a method to build a house like that.

    Now you are a helpful assistant that writes Mineflayer javascript code to complete any Minecraft task specified by me.
    
    Here are some useful programs written with Mineflayer APIs:
    
    await bot.pathfinder.goto(goal); // A very useful function. This function may change your main-hand equipment.
    // Following are some Goals you can use:
    new GoalNear(x, y, z, range); // Move the bot to a block within the specified range of the specified block. `x`, `y`, `z`, and `range` are `number`
    new GoalXZ(x, z); // Useful for long-range goals that don't have a specific Y level. `x` and `z` are `number`
    new GoalGetToBlock(x, y, z); // Not get into the block, but get directly adjacent to it. Useful for fishing, farming, filling bucket, and beds. `x`, `y`, and `z` are `number`
    new GoalFollow(entity, range); // Follow the specified entity within the specified range. `entity` is `Entity`, `range` is `number`
    new GoalPlaceBlock(position, bot.world, {}); // Position the bot in order to place a block. `position` is `Vec3`
    new GoalLookAtBlock(position, bot.world, {}); // Path into a position where a blockface of the block at position is visible. `position` is `Vec3`
    
    // These are other Mineflayer functions you can use:
    bot.isABed(bedBlock); // Return true if `bedBlock` is a bed
    bot.blockAt(position); // Return the block at `position`. `position` is `Vec3`
    
    // These are other Mineflayer async functions you can use:
    await bot.equip(item, destination); // Equip the item in the specified destination. `item` is `Item`, `destination` can only be "hand", "head", "torso", "legs", "feet", "off-hand"
    await bot.consume(); // Consume the item in the bot's hand. You must equip the item to consume first. Useful for eating food, drinking potions, etc.
    await bot.fish(); // Let bot fish. Before calling this function, you must first get to a water block and then equip a fishing rod. The bot will automatically stop fishing when it catches a fish
    await bot.sleep(bedBlock); // Sleep until sunrise. You must get to a bed block first
    await bot.activateBlock(block); // This is the same as right-clicking a block in the game. Useful for buttons, doors, etc. You must get to the block first
    await bot.lookAt(position); // Look at the specified position. You must go near the position before you look at it. To fill bucket with water, you must lookAt first. `position` is `Vec3`
    await bot.activateItem(); // This is the same as right-clicking to use the item in the bot's hand. Useful for using buckets, etc. You must equip the item to activate first
    await bot.useOn(entity); // This is the same as right-clicking an entity in the game. Useful for shearing sheep, equipping harnesses, etc. You must get to the entity first
    
    
    
    At each round of conversation, I will give you
    Nearby blocks: ...
    Position: ...
    Task: ...
    Context: ...
    
    
    You should then respond to me with
    Explain (if applicable): Are there any steps missing in your plan? Why does the code not complete the task? What does the chat log and execution error imply?
    Plan: How to complete the task step by step. You should pay attention to Inventory since it tells what you have. The task completeness check is also based on your final inventory.
    Code:
        1) Write an async function taking the bot as the only argument.
        2) Reuse the above useful programs as much as possible.
            - Use `mineBlock(bot, name, count)` to collect blocks. Do not use `bot.dig` directly.
            - Use craftItem(bot, name, count) to craft items. Do not use bot.craft or bot.recipesFor directly.
            - Use `smeltItem(bot, name count)` to smelt items. Do not use `bot.openFurnace` directly.
            - Use `placeItem(bot, name, position)` to place blocks. Do not use `bot.placeBlock` directly.
            - Use `killMob(bot, name, timeout)` to kill mobs. Do not use `bot.attack` directly.
        3) Your function will be reused for building more complex functions. Therefore, you should make it generic and reusable. You should not make strong assumption about the inventory (as it may be changed at a later time), and therefore you should always check whether you have the required items before using them. If not, you should first collect the required items and reuse the above useful programs.
        4) Anything defined outside a function will be ignored, define all your variables inside your functions.
        5) Call `bot.chat` to show the intermediate progress.
        6) Do not write infinite loops or recursive functions.
        7) Do not use `bot.on` or `bot.once` to register event listeners. You definitely do not need them.
        8) Name your function in a meaningful way (can infer the task from the name).
    
    You should only respond in the format as described below:
    RESPONSE FORMAT:
    Explain: ...
    Plan:
    1) ...
    2) ...
    3) ...
    ...
    Code:
    javascript
    // helper functions (only if needed, try to avoid them)
    ...
    // main function after the helper functions
    async function yourMainFunctionName(bot) {
      // ...
    }
    
    
    Now I will give you information:
    
    Nearby blocks: dirt, grass_block
    
    Position: x=16.5, y=-60.0, z=-127.5
    
    Task: build a house
    
    Context: Build a house according to the figure. Your building should be similar to the one in the image.
    
    
    Here is an example of java script code:
    Code Example:
    javascript
    // helper function to build a house
    async function buildHouse(bot, position, size, blockName) {
        for (let y = 0; y < size; y++) {
            for (let x = 0; x < size; x++) {
                for (let z = 0; z < size; z++) {
                    const targetPosition = position.offset(x, y, z);
                    await placeItem(bot, blockName, targetPosition);
                }
            }
        }
        bot.chat("House built.");
    }
    
    // main function
    async function buildWoodenHouse(bot) {
        const position = bot.entity.position.offset(1, 0, 1); // offset to avoid building at the bot's position
        const size = 5; // size of the house
        const blockName = 'oak_planks'; // material to build the house
        await buildHouse(bot, position, size, blockName);
    }
    
    
    Please note that:
    1) You should not use only one for-loop. Different walls should be built by different for-loops.
    2) Never check whether you have enough blocks in inventory. I will garantee that you will be given enough blocks.
    3) Always use const position = bot.entity.position.offset(1, 0, 1); // offset to avoid building at the bot's position.
    4) Never define placeItem(bot, blockName, targetPosition) by yourself. We already provide a defined function.
    5) Always use const targetPosition = position.offset(...) before placeItem(bot, blockName, targetPosition).
    4) Additionally, y axis always start from 0 rather than 1 in a for-loop.
    5) In terms of the size of the house, the kind of blocks of your selection and other details, please refer to the image and your answers to those questions above.
    
    Here are the names of the commonly used blocks that you can choose from but not limited to:
    ["ice", "packed_ice", "blue_ice", "beacon", "white_concrete", "quartz_block", "smooth_sandstone", "sandstone", "sandstone_slab", "sandstone_stairs", "oak_door", "polished_andesite", "glass", "glass_pane", "lantern", "sea_lantern", "glowstone", "blue_glazed_terracotta", "white_glazed_terracotta", "green_glazed_terracotta", "yellow_glazed_terracotta", "red_glazed_terracotta", "lime_glazed_terracotta", "cyan_glazed_terracotta"]
    You should not misspell them in your code.
    
    One last important thing: you should write your code within maximum length of tokens.
\end{lstlisting}

Then, the generated code is passed to Mineflayer for execution in the game. We recursively restart the conversation until the generated code does not throw an exception.

\section{Evaluation Metric}
\label{appendix:metric}

In this section, we present detailed prompts for all the evaluation metrics and the protocol for human evaluation.

\subsection{VLM Evaluation}
We use GPT-4V for the two evaluation methods: 1v1 comparison and scoring.

For 1v1 comparison, the prompts are as follows:

\begin{lstlisting}
    You are a critic with high aesthetic abilities.  I will provide you a text instruction and two buildings created by different agents following this instruction in the game "Minecraft".

    Please evaluate their overall performance according to four aspects:

    1. Correctness. Are the creations consistent with the language instruction?
    2. Complexity. Can the agent create large and complex buildings?
    3. Quality. Do the creations have good visual appearance?
    4. Functionality. Do the created buildings have necessary structures for houses (rooms and entrances)? 

    Tell me which building in the image is better (left or right).

    Instrution: $(INSTRUCTION)
    Image of buildings: $(IMAGE1, IMAGE2)
\end{lstlisting}

For scoring, the prompts are as follows.

\begin{lstlisting}
    You are a critic with high aesthetic abilities.  I will provide you a text instruction and a building created by an agent following this instruction in the game "Minecraft"

    Please evaluate their overall performance according to four aspects:

    1. Correctness. Are the creations consistent with the language instruction?
    2. Complexity. Can the agent create large and complex buildings?
    3. Quality. Do the creations have good visual appearance?
    4. Functionality. Do the created buildings have necessary structures for houses (rooms and entrances)? 

    please evaluate the building with a score (out of 10) on each aspect respectively, then give a total score.

    instruction: $(INSTRUCTION)
    Image of building: $(IMAGE)
\end{lstlisting}

\subsection{Human Evaluation}

\begin{figure*}[htbp]
  \centering
    \includegraphics[width=1\textwidth]{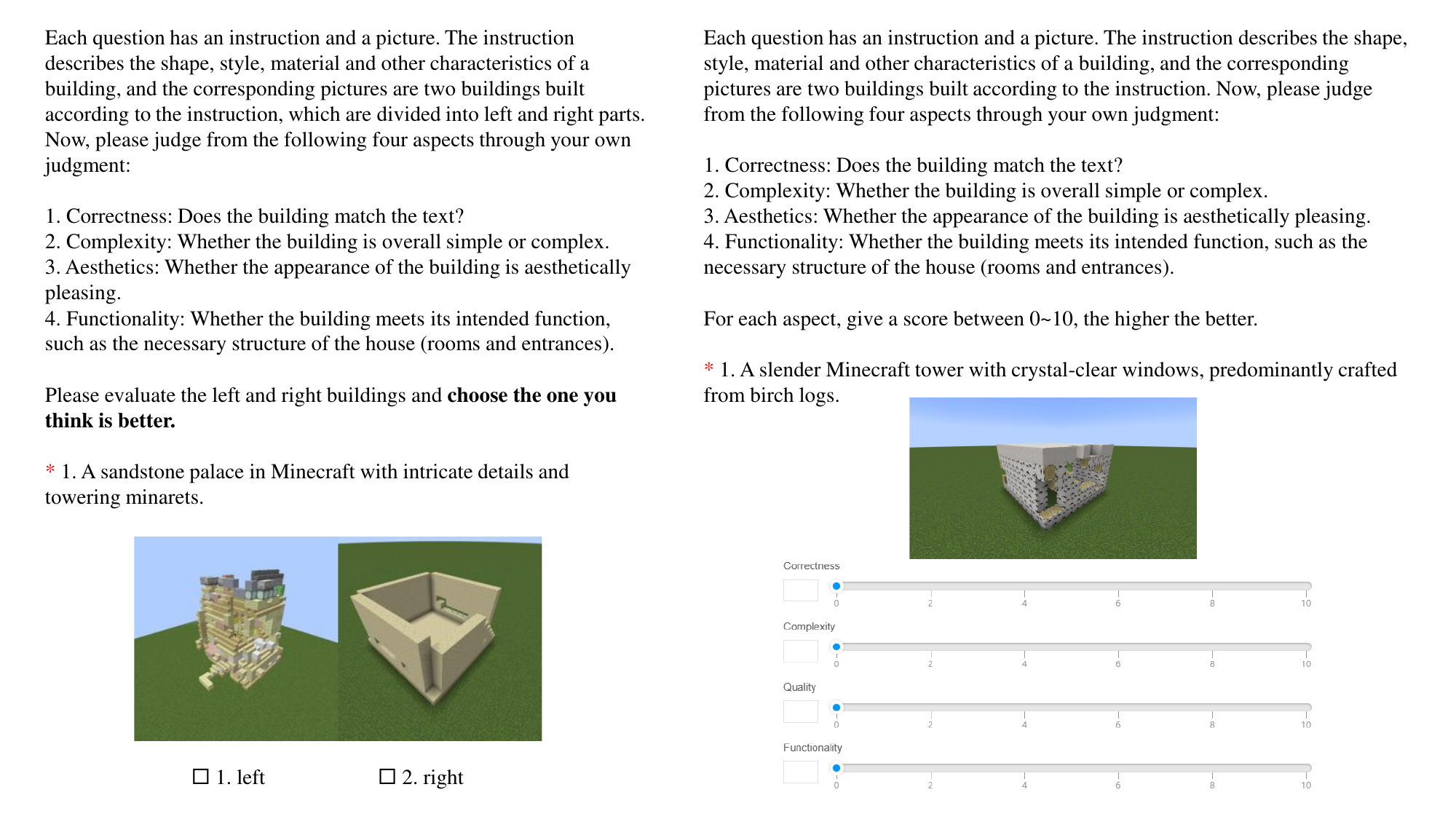}
  \caption{An example of the questionnaires for human evaluation. \textbf{Left}: 1v1 comparison between different methods; \textbf{Right}: directly score the test sample.}
  \label{fig:questionnaire}
\end{figure*}

We conducted the human evaluation in a similar way as the VLM evaluation, while we converted prompts into questionnaires. In total, we received 49 valid questionnaires evaluating the results of the four methods for all the test tasks.
Figure~\ref{fig:questionnaire} demonstrates the questionnaires for both evaluation methods.

\subsection{Linear Regression and Analysis}
\label{linear-regression}

In the scoring process, we collected human ratings and GPT4 scores for the four different methods and four metrics, respectively. Due to data collection constraints, human ratings were divided into four groups for separate assessments. Consequently, in the Ordinary Least Squares (OLS) regression, we incorporated three agent dummy variables to distinguish between the four methods and three group dummy variables to account for the four groups of raters. The results obtained from the OLS linear regression using the aforementioned approach are presented in Table \ref{tab:quality}, \ref{tab:functionality}, \ref{tab:complexity}, \ref{tab:correctness}, \ref{tab:overall}.

\begin{table}
    \centering
    \caption{In the assessment of quality, coefficient, standard errors, t-value, and p-value correspond to each independent variable.}
    \label{tab:quality}
    \setlength{\tabcolsep}{3pt}
    \begin{tabular}{c|cccc}
        \toprule
        \textbf{     Quality     } & \textbf{ coef  } & \textbf{std err} & \textbf{   $t$   } & \textbf{ $P>|t|$ } \\
        \midrule
        const & 5.6957 & 0.423 & 13.463 & 0.001 \\
        
        human & 0.9774 & 0.342 & 2.858 & 0.006 \\
     
        agent1 & -0.4709 & 0.484 & -0.973 & 0.334 \\
    
        agent2 & 0.2293 & 0.492 & 0.466 & 0.643 \\
    
        agent3 & 0.1305 & 0.551 & 0.237 & 0.813 \\
    
        type1 & 0.1497 & 0.462 & 0.324 & 0.747 \\
    
        type2 & 0.1069 & 0.442 & 0.242 & 0.810 \\
    
        type3 & 10.1187 & 0.442 & 0.269 & 0.789 \\
        \bottomrule
    \end{tabular}
\end{table}

\begin{table}
    \centering
    \caption{In the assessment of functionality, coefficient, standard errors, t-value, and p-value correspond to each independent variable.}
     \label{tab:functionality}
    \setlength{\tabcolsep}{3pt}
    \begin{tabular}{c|cccc}
        \toprule
        \textbf{Functionality} & \textbf{ coef  } & \textbf{std err} & \textbf{   $t$   } & \textbf{$ P>|t|$ } \\
        \midrule
        const & 4.5400 & 0.471 & 9.647 & 0.001 \\
        human & 1.0185 & 0.378 & 2.697 & 0.009 \\
        agent1 & -0.4817 & 0.531 & -0.907 & 0.367 \\
        agent2 & -0.1709 & 0.558 & -0.306 & 0.760 \\
        agent3 & -0.5622 & 0.614 & -0.916 & 0.363 \\
        type1 & -0.4929 & 0.507 & -0.971 & 0.335 \\
        type2 & 0.1458 & 0.485 & -0.300 & 0.765 \\
        type3 & -0.5684 & 0.484 & -1.174 & 0.244 \\
        \bottomrule
    \end{tabular}
\end{table}

\begin{table}
    \centering
    \caption{In the assessment of complexity, coefficient, standard errors, t-value, and p-value correspond to each independent variable.}
    \label{tab:complexity}
    \setlength{\tabcolsep}{3pt}
    \begin{tabular}{c|cccc}
        \toprule
        \textbf{ Complexity } & \textbf{ coef  } & \textbf{std err} & \textbf{   $t$   } & \textbf{$ P>|t|$ } \\
        \midrule
        const & 4.5340 & 0.366 & 12.385 & 0.001 \\
        human & 1.9722 & 0.281 & 7.021 & 0.001 \\
        agent1 & -0.0042 & 0.448 & -0.009 & 0.993 \\
        agent2 & 0.5004 & 0.428 & 1.169 & 0.246 \\
        agent3 & -0.1349 & 0.417 & -0.323 & 0.747 \\
        type1 & -0.2672 & 0.424 & -0.630 & 0.531 \\
        type2 & 0.0028 & 0.408 & 0.007 & 0.995 \\
        type3 & -0.1376 & 0.408 & -0.337 & 0.737 \\
        \bottomrule
    \end{tabular}
\end{table}

\begin{table}
    \centering
    \caption{In the assessment of correctness, coefficient, standard errors, t-value, and p-value correspond to each independent variable.}
    \label{tab:correctness}
    \setlength{\tabcolsep}{3pt}
    \begin{tabular}{c|cccc}
        \toprule
        \textbf{Correctness} & \textbf{ coef  } & \textbf{std err} & \textbf{   $t $  } & \textbf{ $P>|t|$ } \\
        \midrule
        const & 5.9786 & 0.585 & 10.227 & 0.001 \\
        human & 1.4794 & 0.404 & 3.663 & 0.001 \\
        agent1 & -0.9468 & 0.660 & -1.435 & 0.156 \\
        agent2 & -0.4817 & 0.691 & -0.698 & 0.488 \\
        agent3 & -0.1575 & 0.723 & -0.218 & 0.828 \\
        type1 & 0.8687 & 0.649 & 1.339 & 0.185 \\
        type2 & 0.5536 & 0.624 & 0.887 & 0.378 \\
        type3 & 0.7495 & 0.623 & 1.204 & 0.233 \\
        \bottomrule
    \end{tabular}
\end{table}

\begin{table}
    \centering
    \caption{In the assessment of overall score, coefficient, standard errors, t-value, and p-value correspond to each independent variable.}
    \label{tab:overall}
    \setlength{\tabcolsep}{3pt}
    \begin{tabular}{c|cccc}
        \toprule
        \textbf{    Overall    } & \textbf{ coef  } & \textbf{std err} & \textbf{   $t$   } & \textbf{$ P>|t|$ } \\
        \midrule
        const & 21.5103 & 1.491 & 14.428 & 0.001 \\
        human & 1.7015 & 0.335 & 5.078 & 0.001 \\
        agent1 & -2.7559 & 1.673 & -1.648 & 0.104 \\
        agent2 & -0.9451 & 1.732 & -0.546 & 0.587 \\
        agent3 & -2.2048 & 1.894 & -1.164 & 0.248 \\
        type1 & 0.5965 & 1.659 & 0.360 & 0.720 \\
        type2 & 0.5053 & 1.598 & 0.316 & 0.753 \\
        type3 & 0.1334 & 1.596 & 0.084 & 0.934 \\
        \bottomrule
    \end{tabular}
\end{table}

Heteroscedasticity in the data can violate the assumptions of the classical linear regression model (MLR) as well as statistical significance. Therefore we conduct Breusch-Pagan test \cite{BPtest} to examine the presence of heteroscedasticity issues. The F-value and p-value for different metrics are presented in Table \ref{tab:BPtest}.

\begin{table}
    \centering
    \caption{Breusch Pagan Test. For data with p-values less than 0.2, we fail to reject the null hypothesis, hence it can be considered that there is no significant heteroscedasticity. There is a higher possibility of heteroscedasticity in the correctness metric. However, the remaining metrics are not affected by heteroscedasticity and do not damage statistical significance.}
    \label{tab:BPtest}
    \setlength{\tabcolsep}{3pt}
    \begin{tabular}{c|cc}
        \toprule
        \textbf{    Metrics    } & \textbf{ F-value  } & \textbf{p-value}  \\
        \midrule
        quality & 2.7215 & 0.0019 \\
        functionality & 1.5435 & 0.1762 \\
        complexity & 1.8602 & 0.0993 \\
        correctness & 1.1991 & 0.3165\\
        overall & 1.8619 & 0.0989 \\
        \bottomrule
    \end{tabular}
\end{table}

\section{Additional Results}
\label{appendix:results}

\begin{figure*}[!t]
  \centering
  \includegraphics[width=.96\textwidth]{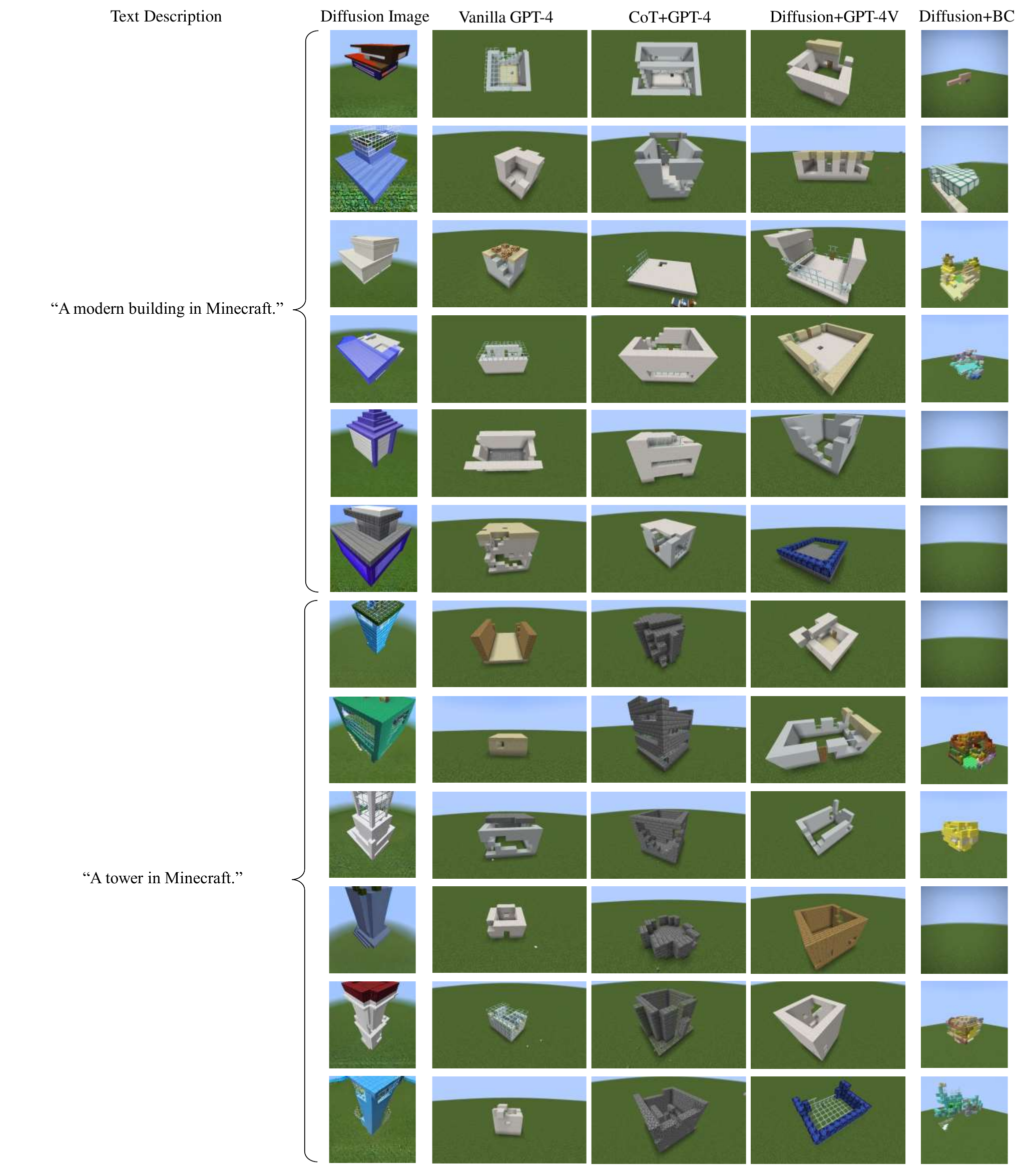}
  \caption{Examples showing the diversity of created buildings, using short language instructions as prompt with minimal information.}
  \label{fig:diversity}
\end{figure*}

The results showing diversity are listed in Figure~\ref{fig:diversity}. We use short language instructions containing little information, and some agents manage to give diverse solutions to these creative tasks.

To study the difference between textual imagination and visual imagination, here we provide an example of the intermediate output after the first round from CoT+GPT-4 and Diffusion+GPT-4V. 
The language instruction of this test task is ``A modern-style building in Minecraft built of quartz block, glass, and lantern.''

The textual imagination from CoT+GPT-4 is:
\begin{lstlisting}
    Designing a modern-style building in Minecraft with a limited inventory of blocks is an interesting challenge. Here's a detailed description based on your specifications:
    
    Description of the Building:
    -	Style: The building is designed in a contemporary, minimalist style, emphasizing clean lines and an open layout.
    -	Exterior: The facade is predominantly made of quartz blocks, creating a sleek, white appearance. Large glass panels are used to bring in natural light and offer views of the surrounding landscape.
    -	Interior: The interior features an open-plan design, with minimal partitions to create a sense of space and fluidity. Lanterns are strategically placed for ambient lighting.
    
    Materials Used:
    1.	Quartz Block: Primarily used for the exterior walls and some interior elements.
    2.	Glass: Used for windows and in some areas for floor-to-ceiling glass walls.
    3.	Lanterns: Used for lighting both indoors and outdoors.
    
    Probable Size of the Building:
    -	Length: Approximately 10 blocks
    -	Width: Approximately 10 blocks
    -	Height: Approximately 10 blocks
    This size ensures that the total number of blocks used remains within the limit of 2304, accounting for the use of different materials.
    
    Components of the Building:
    -	Doors: Modern, minimalist doors made primarily of glass with quartz frames.
    -	Walls: Mainly quartz blocks, with large sections of glass walls.
    -	Windows: Large, glass panels, some of which extend from floor to ceiling.
    -	Floors: Quartz blocks with some areas featuring glass flooring for a contemporary feel.
    -	Roof: A flat roof made of quartz blocks, incorporating a few glass elements for natural lighting.
    
    Salient Features:
    -	Garden Area: A small zen-style garden at the rear, visible through the glass walls, adding a touch of greenery.
    -	Terrace: A rooftop terrace accessible via a staircase, providing a panoramic view of the surroundings.
    -	Swimming Pool: A small, elegant infinity pool on the rooftop, lined with quartz blocks.
    -	Interior Courtyard: A small, open-air courtyard in the center of the building, bringing in natural light and air.
    -	Lighting: Lanterns are used both inside and outside to create a warm, welcoming ambiance during the night.
    
    This building, with its modern design and thoughtful use of materials, would stand out in a Minecraft world, providing both functionality and aesthetic appeal.
\end{lstlisting}

\begin{figure}[h]
  \centering
  \includegraphics[scale=0.32]{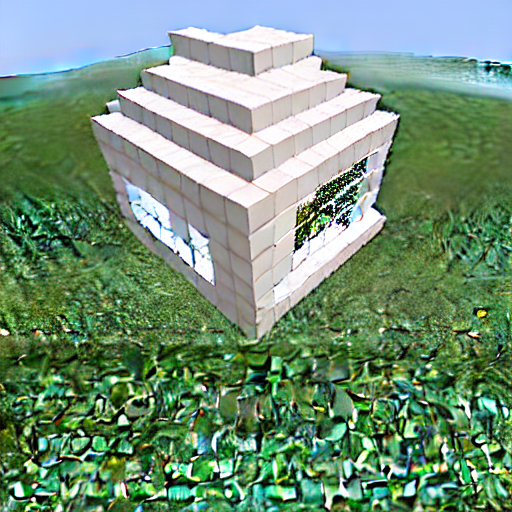}
  \caption{The diffusion visual imagination of the task ``A modern-style building in Minecraft built of quartz block, glass, and lantern.''}
  \label{fig:3_in_appendix}
\end{figure}

The diffusion visual imagination of this task is shown in Figure~\ref{fig:3_in_appendix}.
Given this imagination, the intermediate reasoning results of Diffusion+GPT-4V are:
\begin{lstlisting}
    The building in the image appears to be a modern Minecraft structure with the following characteristics:
    
    1.	Detailed Description: The building has a distinctive step pyramid or ziggurat shape with a tiered structure, each level smaller than the one below it, creating a terraced effect. The design is minimalist and geometric, featuring clean lines and a symmetrical layout. It sits on what seems to be a grassy terrain with a clear sky above, suggesting an open and possibly peaceful setting.
    
    2.	Blocks Used:
    -	Quartz Blocks: These are the predominant material and give the building its white and clean look. They are used for the main structure.
    -	Glass Blocks/Panes: These are used for windows and possibly for some decorative elements, allowing for natural light to enter the building and providing a transparent look.
    -	Leaves: There seems to be foliage incorporated into the building, likely for decorative purposes, suggesting the use of leaf blocks.
    -	Lanterns: If there are sources of light visible, they might be lantern blocks, used to illuminate the structure.
    
    3.	Probable Size:
    -	Length and Width: It's difficult to give exact measurements since the image perspective doesn't show the full base. However, if we assume each block is 1 meter, as is standard in Minecraft, we can estimate based on the visible tiers. If each tier is approximately 5 blocks less in length and width than the one below, and there are 4 visible tiers, the base could be around 20 blocks wide and long.
    -	Height: The building appears to have at least 4 levels. If each level is around 3 blocks in height, the total height could be around 12 blocks.
    
    4.	Components:
    -	Doors: There is no visible door from this angle, but there could be an entrance on a side not shown.
    -	Walls: The walls are primarily made of quartz blocks, with glass blocks likely used for windows.
    -	Windows: Large, square windows can be seen, made out of glass blocks or panes.
    -	Floors: Each level acts as a floor for the structure, likely made of quartz blocks as well.
    
    5.	Salient Features:
    No gardens, swimming pools, or towers are visible in this image. The most salient feature is the building's model architectural style with its stepped, pyramid-like structure. There could be additional features on the other sides of the building not visible in this image. 
    
    Without seeing all sides of the building, this description is based on the visible elements in the provided image.
\end{lstlisting}

\noindent By comparing the two imaginations, we have the following observations:
\begin{itemize}
    \item In terms of the blocks and materials used to build houses, CoT+GPT-4 usually captures block types from the raw input (text description of the building task), while Diffusion+GPT-4V tends to identify blocks and materials based on the images generated by the diffusion model. This also means that the error from the Diffusion-based imaginator will be inherited and amplified, sometimes resulting in GPT-4V's wrong recognition.
    
    \item In terms of conventional structures and components of the building, CoT+GPT-4 can always design such structures including doors, windows, and roofs. However, Diffusion+GPT-4V answers truthfully based on the diffusion images, making a difference from CoT+GPT-4. For instance, if there are only windows appearing in the image, GPT-4V will not respond with structures like doors.

    \item When it comes to other salient features, CoT+GPT-4 can give an imagination with descriptions of gardens, towers, and swimming pools, though these are never implemented in the code generated in the second round. In contrast, Diffusion+GPT-4V denies the existence of such features, since it answers truthfully based on the diffusion images.
\end{itemize}

\end{document}